\documentclass{article}
\usepackage{ijcai20}

\usepackage{times}
\usepackage{soul}
\usepackage{url}
\usepackage[hidelinks]{hyperref}
\usepackage[utf8]{inputenc}
\usepackage[small]{caption}
\usepackage{graphicx}
\usepackage{amsmath}
\usepackage{amsthm}
\usepackage{amsfonts}
\usepackage{booktabs}
\usepackage{algorithm}
\usepackage{algorithmicx}
\usepackage{algpseudocode}
\usepackage{subcaption}
\usepackage[margin=1in]{geometry}
\DeclareMathOperator{\E}{\mathbb{E}}

\title{Learning to Complement Humans}

\author{
Bryan Wilder$^1$\footnote{Research performed during an internship at Microsoft Research.}
\and
Eric Horvitz$^2$ \And
Ece Kamar$^2$\\
\affiliations
$^1$ School of Engineering and Applied Sciences, Harvard University\\
$^2$ Microsoft Research\\
bwilder@g.harvard.edu, horvitz@microsoft.com, eckamar@microsoft.com
}

\usepackage[usenames,dvipsnames]{xcolor}

\begin{document}

\maketitle

\begin{abstract}
    A rising vision for AI in the open world centers on the development of systems that can complement humans for perceptual, diagnostic, and reasoning tasks. To date, systems aimed at complementing the skills of people have employed models trained to be as accurate as possible in isolation. We demonstrate how an end-to-end learning strategy can be harnessed to optimize the combined performance of human-machine teams by considering the distinct abilities of people and machines. The goal is to focus machine learning on problem instances that are difficult for humans, while recognizing instances that are difficult for the machine and seeking human input on them. We demonstrate in two real-world domains (scientific discovery and medical diagnosis) that human-machine teams built via these methods outperform the individual performance of machines and people. We then analyze conditions under which this complementarity is strongest, and which training methods amplify it. Taken together, our work provides the first systematic investigation of how machine learning systems can be trained to complement human reasoning. 
\end{abstract}

\section{Introduction}

 Systems developed via machine learning (ML) are increasingly competent at performing tasks that have traditionally required human expertise, with emerging applications in medicine, law, transportation, scientific discovery, and other disciplines (e.g., \cite{esteva2017dermatologist,chen2018rise,mcginnis2019great}). To date, engineers have constructed models by optimizing model performance in isolation rather than seeking richer optimizations that consider human-machine teamwork. 
 
 Optimizing ML performance in isolation overlooks the common situation where human expertise can contribute complementary perspectives, despite humans having their own limitations, including systematic biases \cite{tversky1974judgment}. We introduce methods for optimizing team performance, where machines take on some parts of the task and humans others. In an ideal world, the machine would be able to handle all instances itself. For complex domains though, this rarely holds in practice, whether due to limited data or model capacity, outliers, superior perceptual or reasoning abilities of people on a given task, or evidence or context available only to humans. When perfect accuracy is unattainable, the machine should focus its limited capacity on regions of the space where it offers the most benefit  (e.g., on cases that are challenging for humans), while pursuing human expertise to handle others. We develop methods aimed at training the ML model to complement the strengths of the human, accounting for the cost of querying an expert. While human-machine teamwork can take many forms, we focus here on settings where a machine takes on the tasks of deciding which instances require human input and then fusing machine and human judgments.   

\begin{figure}
    \centering
    \includegraphics[width=3.3in]{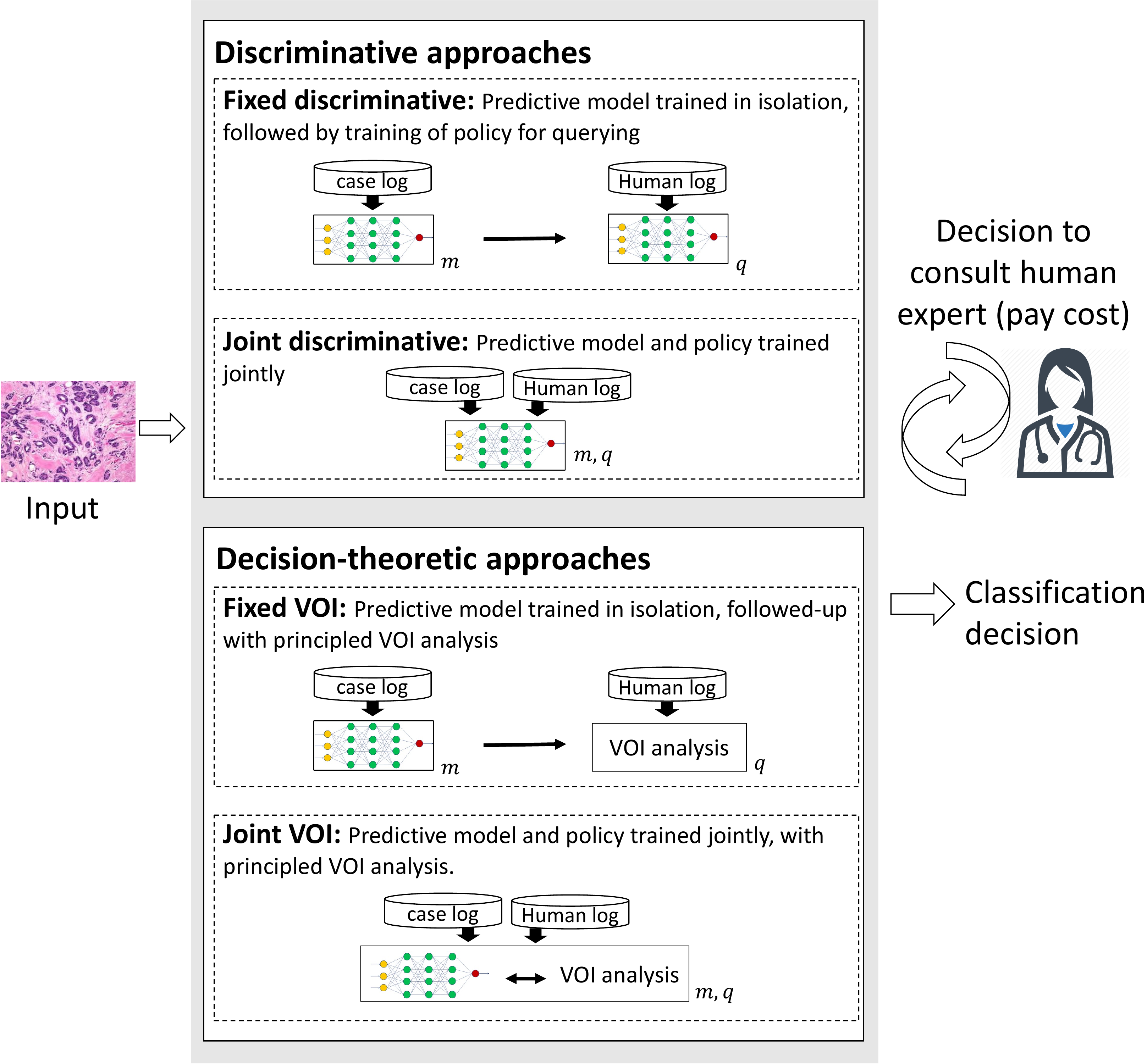}
    \caption{Illustration of task and proposed approaches.} \label{fig:task}
\end{figure}


Prior work includes systems that determine when to consult humans \cite{horvitz2007complementary,kamar2012combining,raghu2019algorithmic}. However, the predictive models are still trained to maximize their own, solitary performance, rather than to leverage the distinctive strengths of machines and humans. The latter requires a shift in the learning objective so as to optimize team performance via instance-sensitive decisions about when to seek human input. To our knowledge, the methods we present are the first to optimize human-AI teams by jointly training ML systems together with policies for allocating tasks to human experts versus machines. We make four contributions:


First, we propose a family of approaches to training an ML system for human-machine complementarity as schematized in Figure \ref{fig:task}. The run-time system combines machine predictions with human input, which may come at additional cost. During training, we use logged human responses to the task to simulate queries to a human. We study both \emph{discriminative} and \emph{decision-theoretic} approaches to optimizing model performance, taking the complementarity of humans and machines into consideration. A baseline approach in either family would first construct an ML model to predict the answer to a given task and then build a policy for deciding when to query the human, taking the predictive model as fixed. We introduce the first generic procedures that operate end-to-end, focused on team performance. With these approaches, we jointly optimize the predictive model and the query policy for team performance, accounting for human-machine complementarities. In the discriminative setting, we introduce a combined loss function that uses a soft relaxation of the query policy for training, along with a technique for making discrete query decisions at run time. In the decision-theoretic setting, we introduce a differentiable surrogate for value of information (VOI) calculations, which allows joint training of the predictive model and the VOI-based query policy through backpropagation. In both cases, joint training focuses the predictive model on instances where the human will not be queried, amplifying complementarity.

Second, we demonstrate the benefits of optimizing for team performance in human-machine teams for two real-world domains of societal importance: scientific discovery (a galaxy classification task) and medical diagnosis (detection of breast cancer metastasis). Via comparative studies, we highlight the importance of guiding learning to optimize the performance of human-machine teams. 

Third, we pursue experimental insights about when and how complementarity-focused training provides benefits. We find evidence for two conclusions: First, training for complementarity is most important when the ML model has limited capacity, forcing it to pick parts of the task to focus on. This suggests that an emphasis on team performance is particularly necessary for difficult tasks that machines cannot perfectly master on their own. Second, training for complementarity has larger benefits when there is an asymmetric cost to errors (e.g., false negatives are more costly than false positives). The need to prioritize among potential errors increases the returns of optimizing for team utility. 

Fourth, we analyze how our methods distribute instances to the human and machine and how these allocations reflect differences in relative capabilities. We find that humans and machines may make qualitatively different kinds of errors. Moreover, the errors made by the ML model change under joint training as the model places more emphasis on instances that are difficult for humans. Via joint training, human and machine errors become different in structured ways that can be leveraged by the methods to improve team performance.


\subsection*{Related Work} Previous work shows that human-machine teams can be more effective than either individually \cite{horvitz2007complementary,kamar2012combining}, including for medical domains \cite{wang2016deep,raghu2019algorithmic}. However, in some others \cite{tan2018investigating,zhang2020effect}, potential complementarity has been difficult to leverage.  
    
Sharing our motivation for developing techniques that harness human-machine complementarity, the work by \cite{raghu2019algorithmic} and \cite{de2020regression} study when a model should outsource a given instance to a human.  \cite{raghu2019algorithmic} is most closely related to our fixed decision-theoretic algorithm; their approach considers predictive variance for the human and machine at each point to allocate human effort. However, the ML model is always fixed, instead of being trained for complementarity. \cite{de2020regression} propose a method to select the parameters of a ridge regression model jointly with a set of training instances to allocate to the human. Our work differs in three important ways: (i) they do not train a query policy to allocate new instances at run time, (ii) our methods apply to arbitrary differentiable models (not just ridge regression), (iii) we provide a characterization of why some methods are more or less effective at leveraging complementarity. 
    
Other related work addresses the complementary question of designing ML models as an aid for a human who is charged with making decisions \cite{grgic2019human,green2019principles,hilgard2019learning,lage2018human}. Some of this work emphasizes the need for ML models to account for human reasoning, in particular for humans to learn when to trust the ML model \cite{bansal2019beyond,bansal2019updates}, but does not optimize the model for complementarity. We focus on cases were the ML system decides which instances require human input. 

\section{Problem Formulation}

We formalize the problem of optimizing human-AI teamwork for predictive tasks. We start with the standard supervised learning setting, predicting labels $y \in \mathcal{Y}$ from features $x \in \mathcal{X}$. We focus on multiclass classification, where $\mathcal{Y}$ is a discrete set, but our methods apply to regression with minor modifications. As is typical, we train a model $m$ with parameters $\theta$, which produces a prediction $\hat{y} = m_\theta(x)$. The difference is that each instance may also be labeled by a human. Our training data contains instances $\{(x, y, h)\}_1^N \sim P$ where $h \in \mathcal{Y}$ is a human's prediction and $P$ is an (unknown) joint distribution. The machine must decide, for each instance, whether to predict on its own or first consult a human expert.

Specifically, the machine learning model first sees $x$ and then decides whether to pay a cost $c$ to observe $h$. $q_\theta(x)$ denotes the query policy, which outputs 1 when the human is queried and 0 otherwise. The model makes a prediction $\hat{y}$, which may depend on $h$ if $q_\theta(x) = 1$. The team's utility is $u(y, \hat{y})$ if the human is not queried, and $u(y, \hat{y}) - c$ if they are. One choice for the utility is $u(y, \hat{y}) = 1[y = \hat{y}]$ (predictive accuracy), but our framework extends easily to asymmetric weightings of different errors. We aim to maximize out-of-sample utility,

\begin{align} \label{eq:objective}
    \E_{(x,y,h) \sim P}\Big[q_\theta(x)\left(u(y, m_\theta(x, h)) - c\right) \\+ (1-q_\theta(x))\left(u(y, m_\theta(x)) \right)\Big]\nonumber.
\end{align}

The first term gives the team utility when the human is queried, and the second when they are not. Conventional supervised learning targets only the second term; our formulation includes the query decision, and the impact of the additional information provided by the human, on the team's overall accuracy.

\section{Approach}

A standard approach to optimizing for human-machine teamwork would first train the model in isolation $m$ to predict the labels $y$ given $x$. Then, $m$ is taken as fixed when constructing the query policy $q$ (as, e.g., in \cite{raghu2019algorithmic}). We propose an alternate approach: joint training that considers explicitly the relative strengths of the human and machine. We introduce methods for both discriminative and decision-theoretic approaches, and now introduce each family in more detail. 

\subsection{Discriminative Approaches}

Discriminative approaches learn functions for $m$ and $q$ which directly map from features to decisions, without building intermediate probabilistic models for the different components of the system. We first introduce a baseline ``fixed" method for training a discriminative system and then propose a means to jointly train the model and query policy together for complementarity with people. 

\subsubsection{Fixed Discriminative Approach}

Traditional \emph{fixed discriminative} approaches train a model $m$ in isolation to perform the task, making the assumption that there is no ability to query the human. That is, we train $m$ to optimize $\E_{(x,y) \sim P}[u(y, m_\theta(x))]$ using any number of well-established methods. Then, taking $m$ as fixed, we construct a query policy $q$ by optimizing Equation \ref{eq:objective}. 

\subsubsection{Joint Discriminative Approach}

In distinction to the fixed approach, we present a \emph{joint discriminative} method that trains the ML model $m_\theta$ end-to-end with the query policy $q_\theta$ so that $m_\theta$ can prioritize instances allocated by $q_\theta$ to the machine. The goal is to optimize a training surrogate for the team utility in Equation \ref{eq:objective}. In the notation, $m_\theta(x)$ denotes the distribution over classes output by the model, and $h$ gives the one-hot encoding of the human responses. 

We propose a differentiable surrogate for Equation \ref{eq:objective}, which can be optimized via stochastic gradient descent whenever the models are themselves differentiable (e.g., neural networks). During training, we will allow $q_\theta(x)$ to take continuous values. This soft relaxation both ensures differentiability and speeds learning by propagating gradient information for both cases (querying and not querying). The most direct relaxation for Equation \ref{eq:objective} is 
\begin{align*}
     q_\theta(x)\ell(y, m_\theta(x,h)) + (1-q_\theta(x))\ell(y, m_\theta(x)) + cq_\theta(x)
\end{align*}
where $\ell$ is any standard loss, which may be weighted to capture asymmetries in the utility $u$. This replaces the potentially discontinuous $u$ with a differentiable loss $\ell$ defined on soft predictions (probability distributions), along with a penalty scaling $c$ by the query probability $q_\theta(x)$. In experiments, this direct relaxation often produced unstable training; intuitively, the predictions and query policy may be spiky in some regions, giving a rapidly changing training signal. The loss we use is
\begin{align*}
     \ell(y, q_\theta(x)m_\theta(x, h) + (1-q_\theta(x))m_\theta(x)) + cq_\theta(x)
\end{align*}
which measures the loss of a fractional prediction that combines the human and machine outputs. The combination tends to behave more smoothly, enabling better training. A key feature of this loss is that it allows the predictions $m_\theta(x)$ to focus on instances that rely heavily on the machine. If $q_\theta(x)$ for some $x$ is close to 1, then the loss for $x$ depends only weakly on $m_\theta(x)$, incentivizing $m$ to focus on instances where $q$ is lower instead. 

When the human is queried, the general formulation allows $m_\theta(x, h)$ to output a prediction different than the human response $h$. However, we observe stronger empirical performance using the simplification $m_\theta(x, h) = h$ (though training a separate model for $m_\theta(x, h)$ results in similar qualitative conclusions). Intuitively, often the correct decision after querying is to output $h$, and including a separate model only adds unnecessary parameters. 

For this simplified formalization, we introduce the following run-time query policy: we need a way of converting the fractional $q$ to a 0 or 1 decision (whether to actually query the human). In an idealized setting where the human label was free, the run-time prediction would be $\arg\max \left(q_\theta(x)h + (1 - q_\theta(x))m_\theta(x)\right)$ (i.e., the highest-probability label in the combined prediction). A naive thresholding scheme would query the human if $q_\theta(x) > 0.5$ (or another fixed value). However, we can approximate the idealized prediction more closely by incorporating a measure of the ML model's confidence, $\max \left(m_\theta(x)\right)$. Specifically, we query the human if
\begin{align*}
    (1 - q_\theta(x)) \max \left(m_\theta(x)\right) < q_\theta(x)
\end{align*}
which results in a query if $q_\theta(x)$ is sufficiently high, \emph{or} the model is sufficiently uncertain. More formally, when this condition holds, the idealized prediction must align with $h$ since $\max \left(q_\theta(x)h\right) > \max\left( (1 - q_\theta(x))m_\theta(x)\right)$. 

\subsection{Decision-Theoretic Approaches} \label{section:voi}

A decision-theoretic approach to human-machine teams, as described in \cite{kamar2012combining}, is to construct probabilistic models for both the ML task and the human response. This allows a follow-up step that calculates the expected value of information for querying the human. 

\subsubsection{Fixed Value of Information Approach}
The \emph{fixed value of information (VOI)} method trains three probabilistic models. $p_\alpha(y|x)$ models the distribution of the label given the features, $p_\beta(h|x)$, the human response given the features, and $p_\gamma(y|h,x)$,  the label given both the features and the human response. $\alpha, \beta, \gamma$ are model parameters. Each model is individually trained to fit its intended target. In our implementation, we use neural networks trained via gradient descent, followed by a sigmoid calibrator trained using the Platt method \cite{platt1999using,niculescu2005predicting}. Calibration is necessary for the predicted probabilities to give meaningful expected utilities.

At execution time, we use these models to estimate the value of querying the human. The estimated expected utility of the ML model without querying the human is

\begin{align*}
    u_{\text{nq}} = \max_{\hat{y} \in \mathcal{Y}} \left(\sum_{y \in \mathcal{Y}} p_\alpha(y|x) u(\hat{y}, y)\right)
\end{align*}

i.e., the value of the prediction with highest expected utility according to $ p_\alpha(y|x)$. Before querying the human, we cannot know the value of $h$ and hence the post-query distribution $p_\gamma(y|x,h)$ is also unknown. However, we can estimate the expected utility by averaging over $p_\beta(h|x)$,

\begin{align*}
u_{\text{q}} =  \E_{h \sim p_\beta(h|x)}\left[ \max_{\hat{y} \in \mathcal{Y}} \left(\sum_{y \in \mathcal{Y}} p_\gamma(y|x,h) u(\hat{y}, y)\right)\right] - c 
\end{align*}

and then query the human whenever $u_{\text{q}} > u_{\text{nq}}$.

\subsubsection{Joint Value of Information Approach} \label{section:e2evoi}

We propose a new decision-theoretic method, which we refer to as  a \emph{joint VOI} approach, that optimizes the utility of the combined system end-to-end, instead of training the best probabilistic model for each individual component. Retaining the structure of the fixed VOI system can be viewed as an inductive bias which allows the model to start from well-founded probabilistic reasoning and then to be fine-tuned for complementarity. To benefit from this inductive bias, we instantiate each of the probabilistic models $p_\alpha, p_\beta$, and $p_\gamma$ with a neural network followed by a Platt calibration layer, just like the fixed VOI approach. However, with joint VOI all of the neural network parameters are trained together via an end-to-end loss, 	\begin{algorithm}
	\caption{Joint VOI training} \label{algorithm:voi}
	\begin{algorithmic}[1] 
		\For{$T$ iterations}
		\State Sample a minibatch $B \subseteq [n]$
		\For{$i \in B$}
		\For{$\hat{y} \in \mathcal{Y}$}
	    \State $u_{\text{nq}}(\hat{y}) = \sum_{y \in \mathcal{Y}} p_\alpha(y|x_i) u(\hat{y}, y)$
	    \EndFor
		\State $u_{\text{nq}} = \sum_{\hat{y} \in \mathcal{Y}} \frac{u_{\text{nq}}(\hat{y}) \exp(u_{\text{nq}}(\hat{y}))}{\sum_{y' \in \mathcal{Y}} \exp(u_{\text{nq}}(y'))}$
		\For{$\hat{y} \in \mathcal{Y}$}
		\State $u_{\text{q}}(\hat{y}, h) = \sum_{y \in \mathcal{Y}} p_\gamma(y|x_i, h) u(\hat{y}, y)$
		\EndFor
		\State $u_{\text{q}} = \sum_{h \in \mathcal{Y}} p_\beta(h|x) \sum_{\hat{y}} \frac{u_{\text{q}}(\hat{y}, h) \exp(u_{\text{q}}(\hat{y}, h))}{\sum_{y' \in \mathcal{Y}} \exp(u_{\text{q}}(y', h))}$
		\State $q = \frac{\exp(u_{\text{q}})}{\exp(u_{\text{q}}) + \exp(u_{\text{nq}})}$
		\State $\ell^i_{\text{combined}} = \ell(q \: p_\gamma(\cdot|x_i, h_i)$
		\State \hspace{\algorithmicindent} $+ (1-q)p_\alpha(\cdot|x_i)) + qc$
		\EndFor
		\State Backpropagate $\frac{1}{|B|}\sum_{i \in B} \ell^i_{\text{combined}}$
		\State Every $t$ iterations: update calibrators
		\EndFor
	\end{algorithmic}
\end{algorithm}which is grounded in the VOI calculation. We update the calibration layer every $t$ steps to maintain well-calibrated probabilities.   

Algorithm \ref{algorithm:voi} outlines joint VOI training. We optimize a surrogate for team utility via stochastic gradient descent, so each iteration first samples a minibatch of data points. For each point, we simulate a differentiable VOI calculation which draws on soft versions of the team's utility if the human were queried ($u_{\text{q}}$) and if the human were not queried ($u_{\text{nq}}$), along with the cost to query. Specifically, line 4 computes $u_{\text{nq}}(\hat{y})$, the expected utility of predicting $\hat{y}$ (according to $p_\alpha$) when the human is not queried. Line 5 takes a softmax over all potential $\hat{y}$ in order to achieve a differentiable approximation to the best achievable expected utility without a query. Similarly, line 6 computes the expected utility $u_{\text{q}}(\hat{y}, h)$ of predicting $\hat{y}$ supposing that the human was queried and responded with $h$. Line 7 takes a softmax over $\hat{y}$ for each fixed $h$ (the inner sum), and then an expectation over $h \sim p_\beta$ (the outer sum). This approximates the expected utility of observing $h$ and then predicting the best $\hat{y}$ given the observation. Line 8 makes a soft query decision via a softmax over $u_{\text{nq}}$ and $u_{\text{q}}$.

Using the output (query decision and prediction) of the differentiable VOI calculation, we compute a team loss $\ell_{\text{combined}}$, which uses the same form as in the joint discriminative model. We average this loss over the minibatch and backpropagate it to update the predictive models.  During this process, we freeze the parameters of the calibration layers of the models. The calibration layers are updated using the Platt procedure every $t$ steps in order to ensure that the model remains well-calibrated even under end-to-end training.

Compared to the fixed model, the joint model uses well-calibrated models to calculate the expected utility of a query. However, it encourages these models to fit most carefully to parts of the space that the are best handled by the machine, and obtains human expertise for others.

\section{Experiments}

We conducted experiments in two real-world domains to explore opportunities for human-machine complementarity and methods to best leverage the complementarity.

\subsection{Domains}

We first explore a scientific discovery task from the Galaxy Zoo project. Here, citizen scientists label images of galaxies as one of five classes to help understand the distribution of galaxies and their evolution. We use 10,000 instances for training and 4,000 for testing. Each instance contains visual features which previous work extracted from the dataset \cite{lintott2008galaxy,kamar2012combining} for $x$. The human response $h$ is the label assigned by a single volunteer (who may make mistakes), while the ground truth $y$ is the consensus over many ($>30$) volunteers.

We next study the medical diagnosis task of detecting breast cancer metastasis in lymph node tissue sections from women with a history of breast cancer. We use data from the CAMELYON16 challenge \cite{bejnordi2017diagnostic}. Each instance contains a whole-slide image of a lymph node section. Each image was labeled by an expert pathologist with unlimited time, providing the ground truth $y$. It was also labeled by a panel of pathologists under realistic time pressure whose diagnoses contain errors; we sample $h$ from the panel responses.


The dataset consists of 127 images. There are also 270 images without panel responses, with which we pretrain the ML models. To develop our models, we follow common practice from high-scoring competition entries (our implementation is based on \cite{vekariyacode}). We first train a convolutional network (Inception-v3 \cite{szegedy2016rethinking}) to predict whether cancer is present in 256$\times$256 pixel patches sampled from the larger whole-slide images. Then, we use Inception-v3 to predict the probability of cancer in each patch, giving a probability heatmap for each slide. We extract visual features from the heatmap (e.g., size of the largest cancer region, eccentricity of the enclosing ellipse, etc). These features are the input $x$ into the human-AI task. This workflow produced the highest-scoring competition entries, ensuring we compare using a state-of-the-art ML method.

\subsection{Models}

We compare each of the four approaches introduced earlier: fixed versus joint discriminative and VOI models. All use neural networks with ReLU activations and dropout ($p = 0.2$). Our experiments vary the number of layers and hidden units to examine the impact of model capacity. We also show a ``Human only" baseline that always queries the human and outputs their response $h$.

    \begin{figure*}
    
    \includegraphics[height=0.9in]{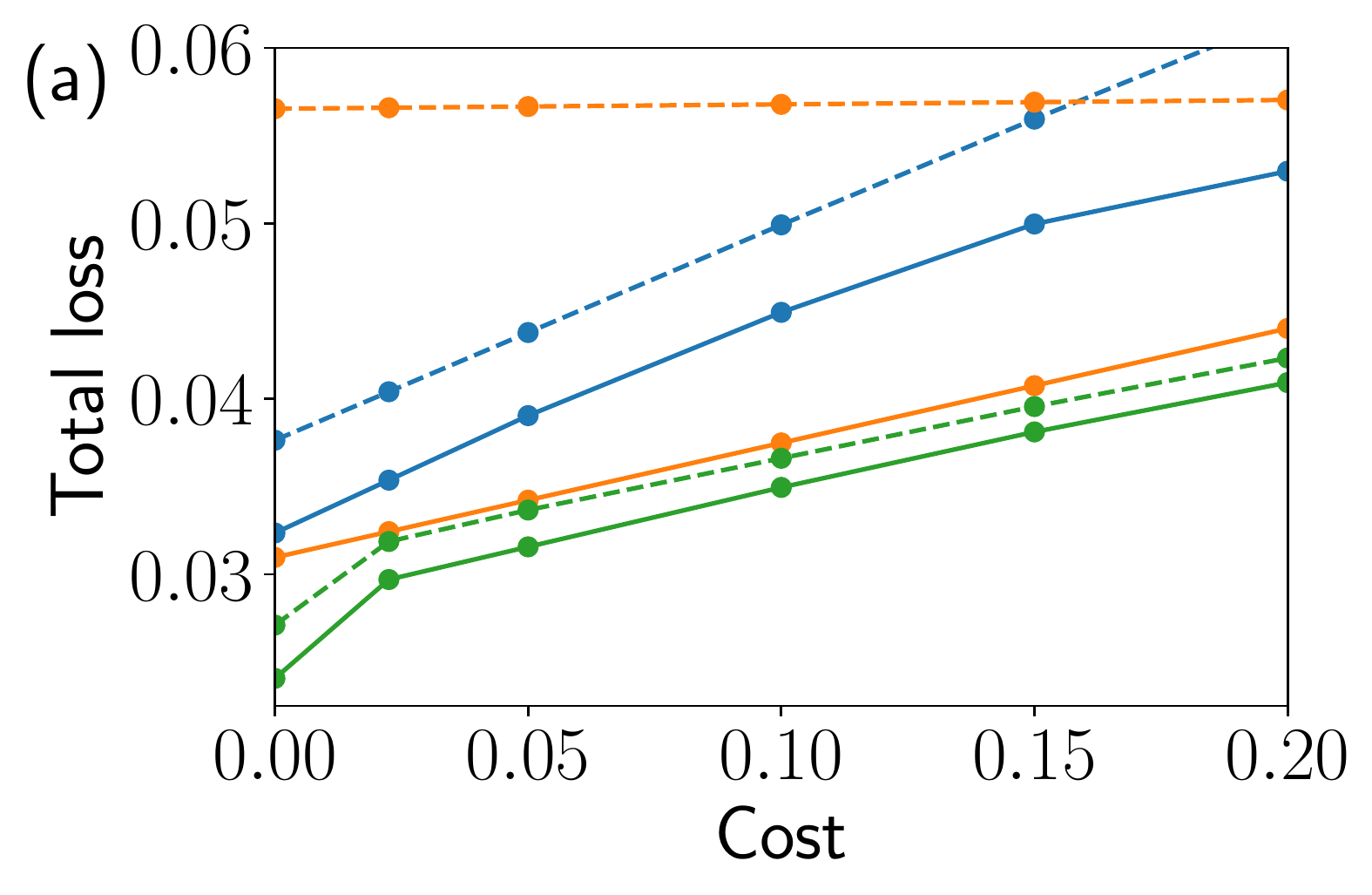}
    \includegraphics[height=0.9in]{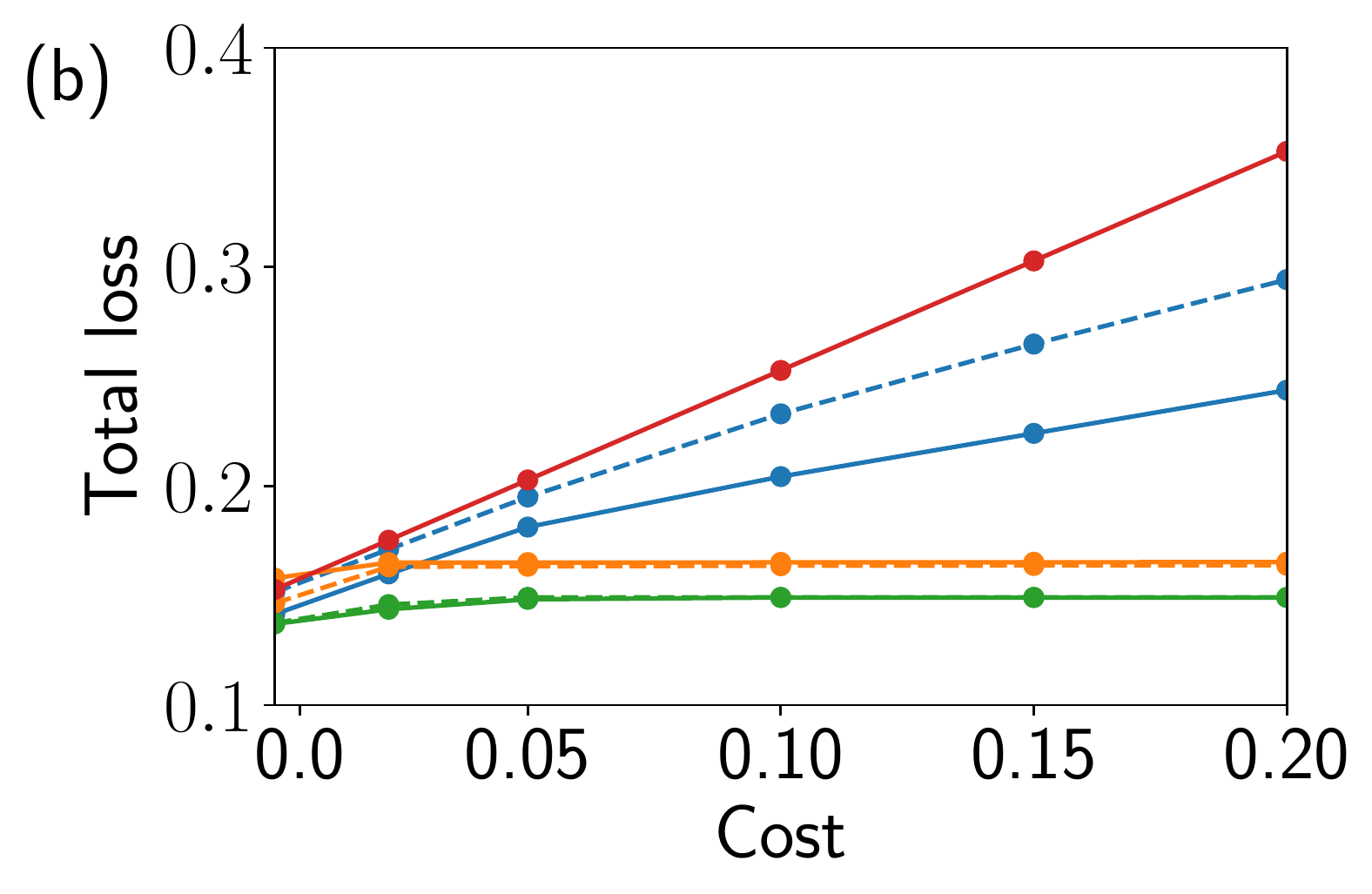}
    \includegraphics[height=0.9in]{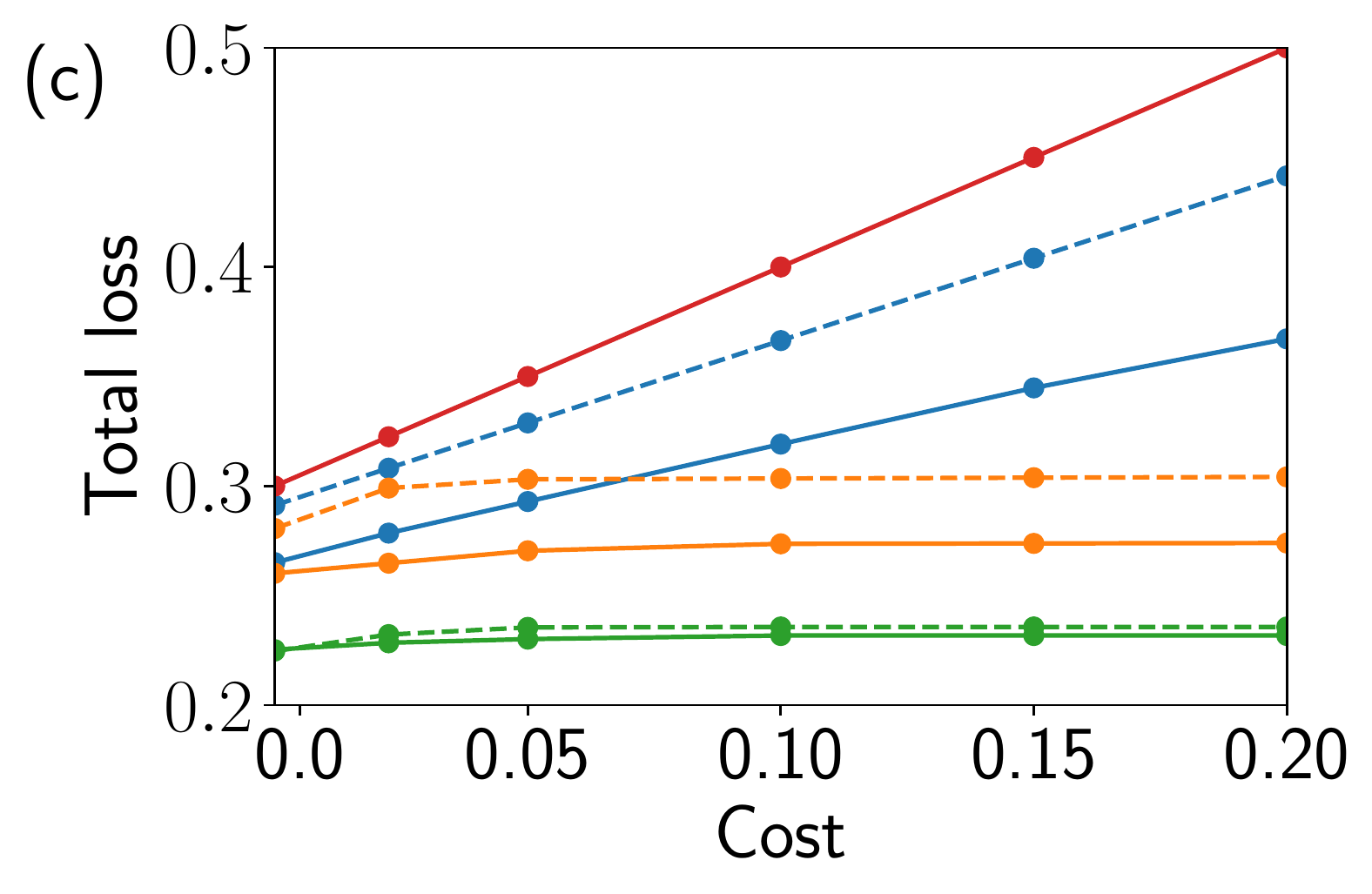}
    \includegraphics[height=0.9in]{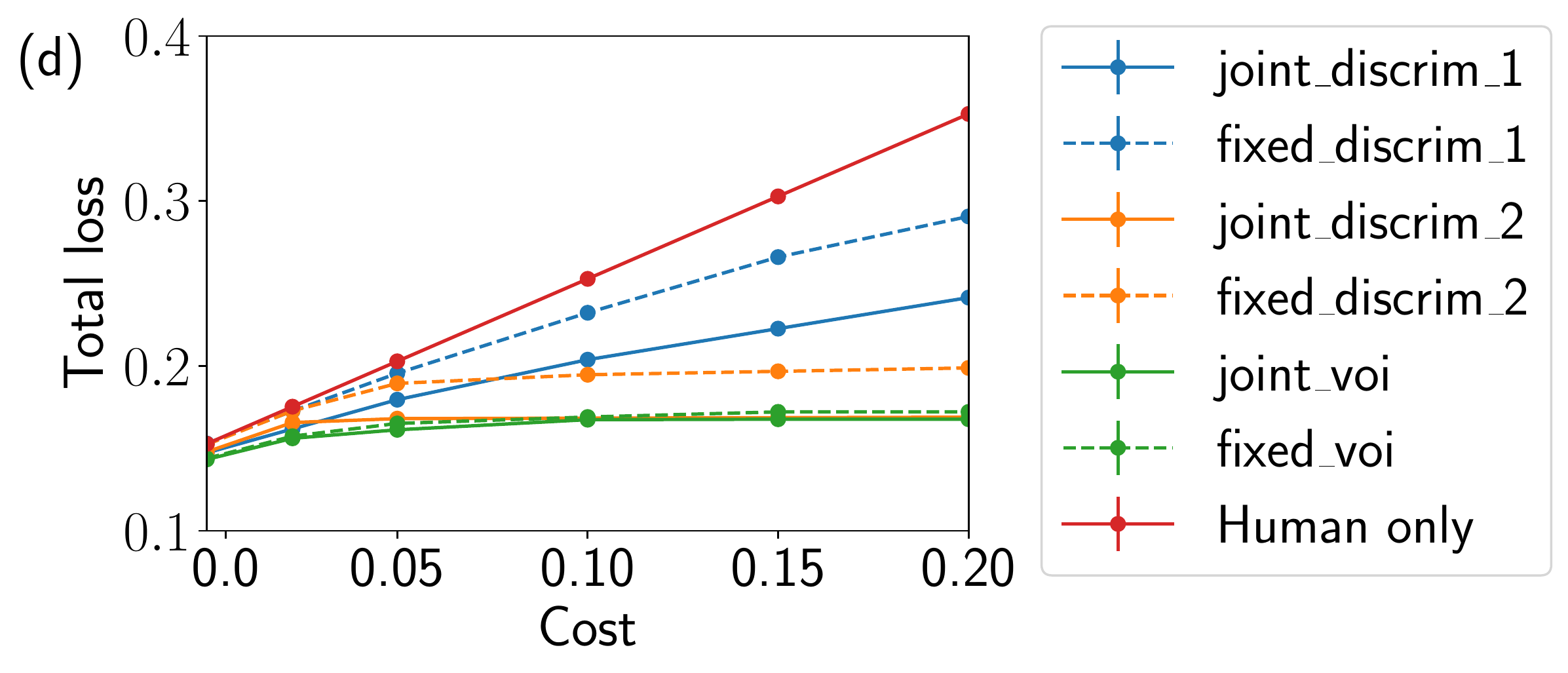}
    
    \includegraphics[height=0.9in]{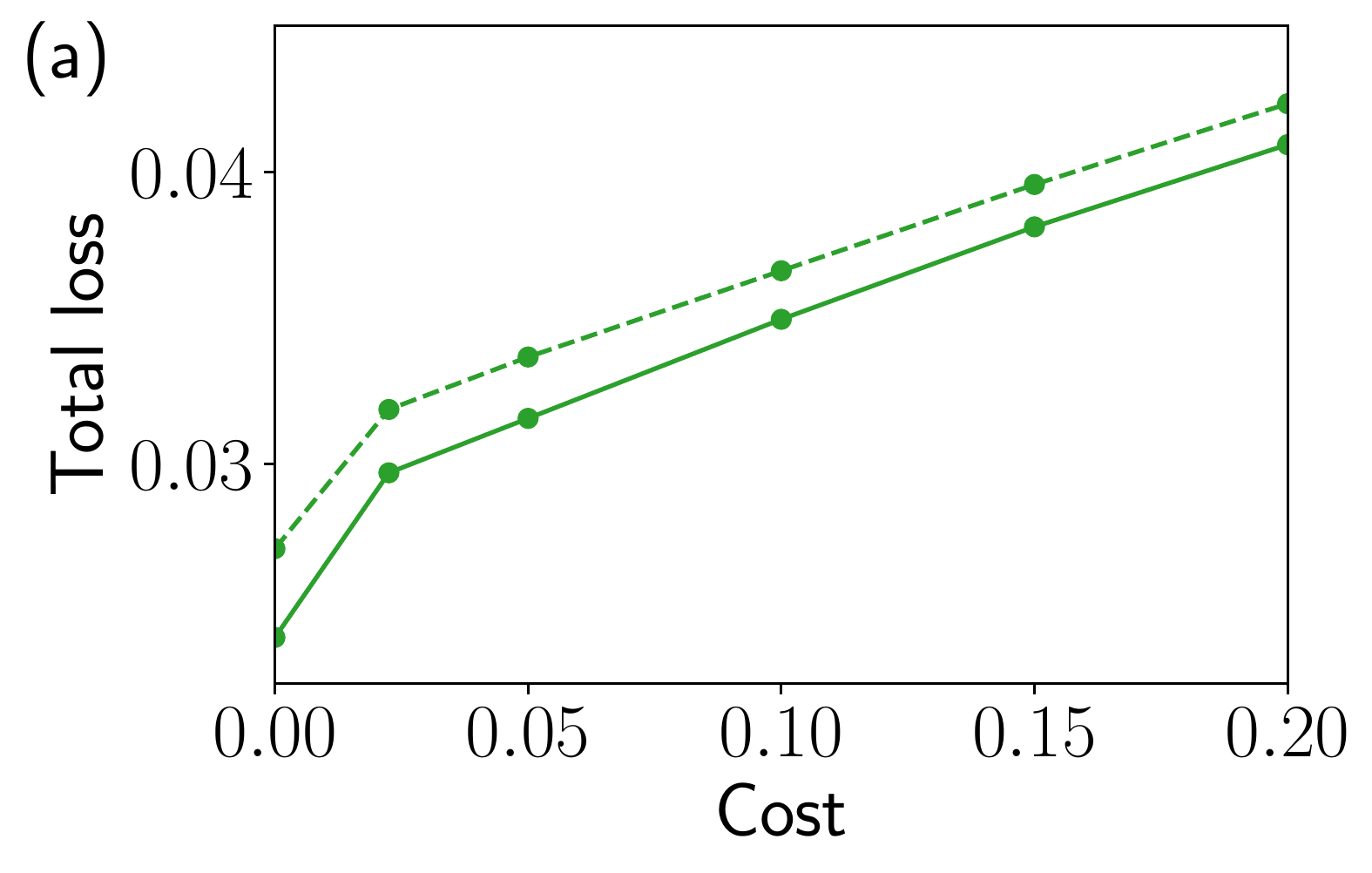}
    \includegraphics[height=0.9in]{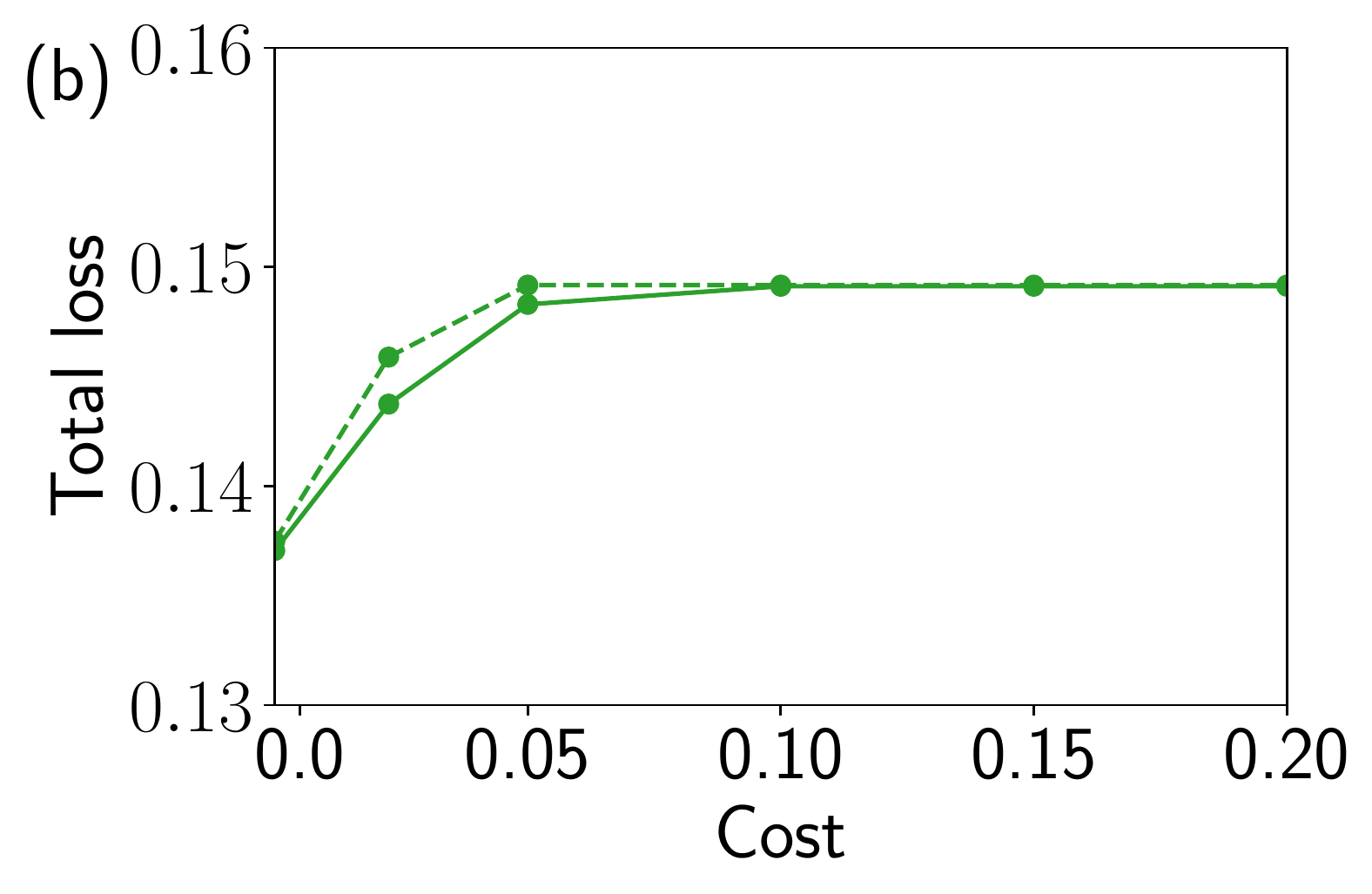}
    \includegraphics[height=0.9in]{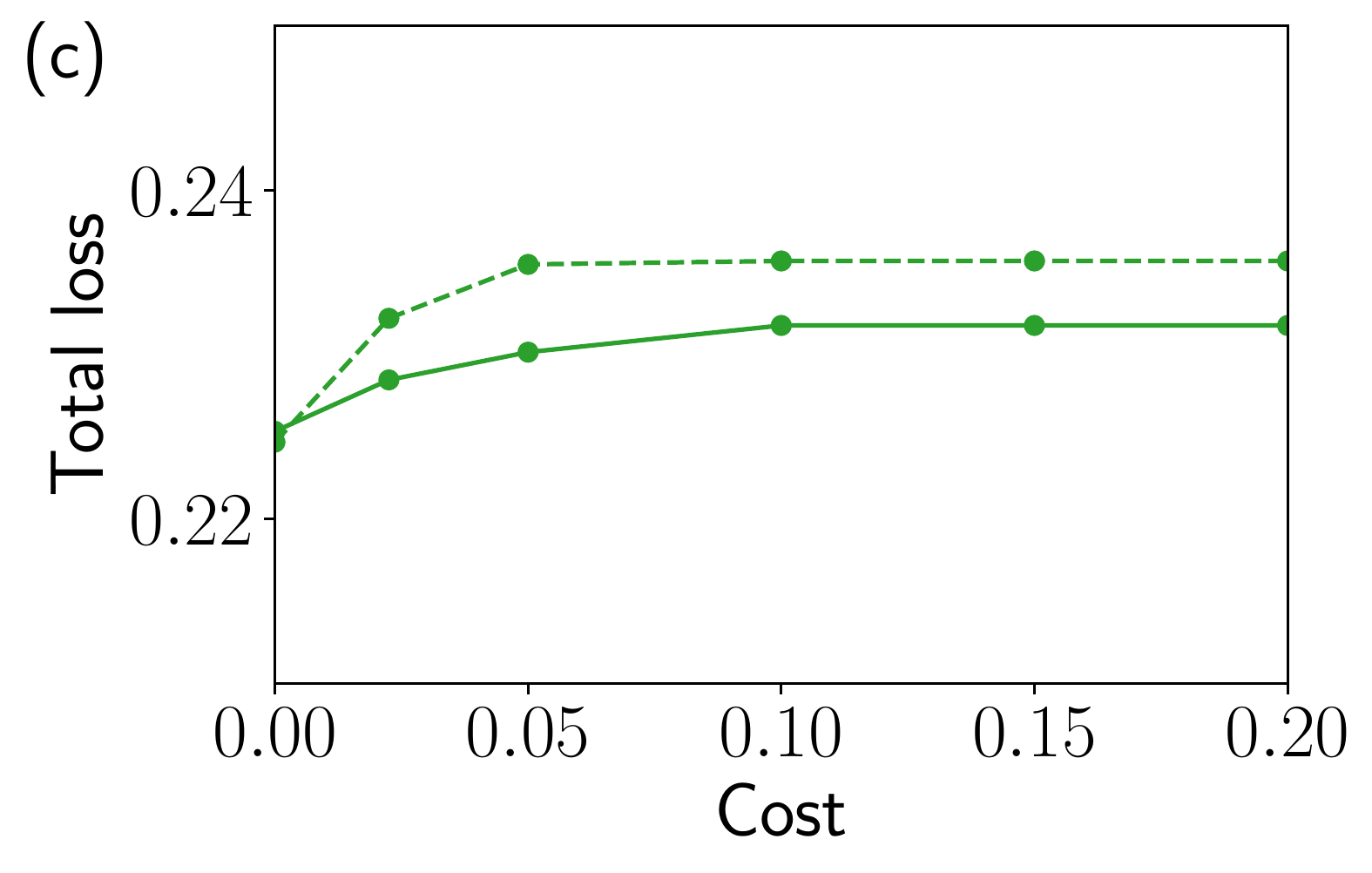}
    \includegraphics[height=0.9in]{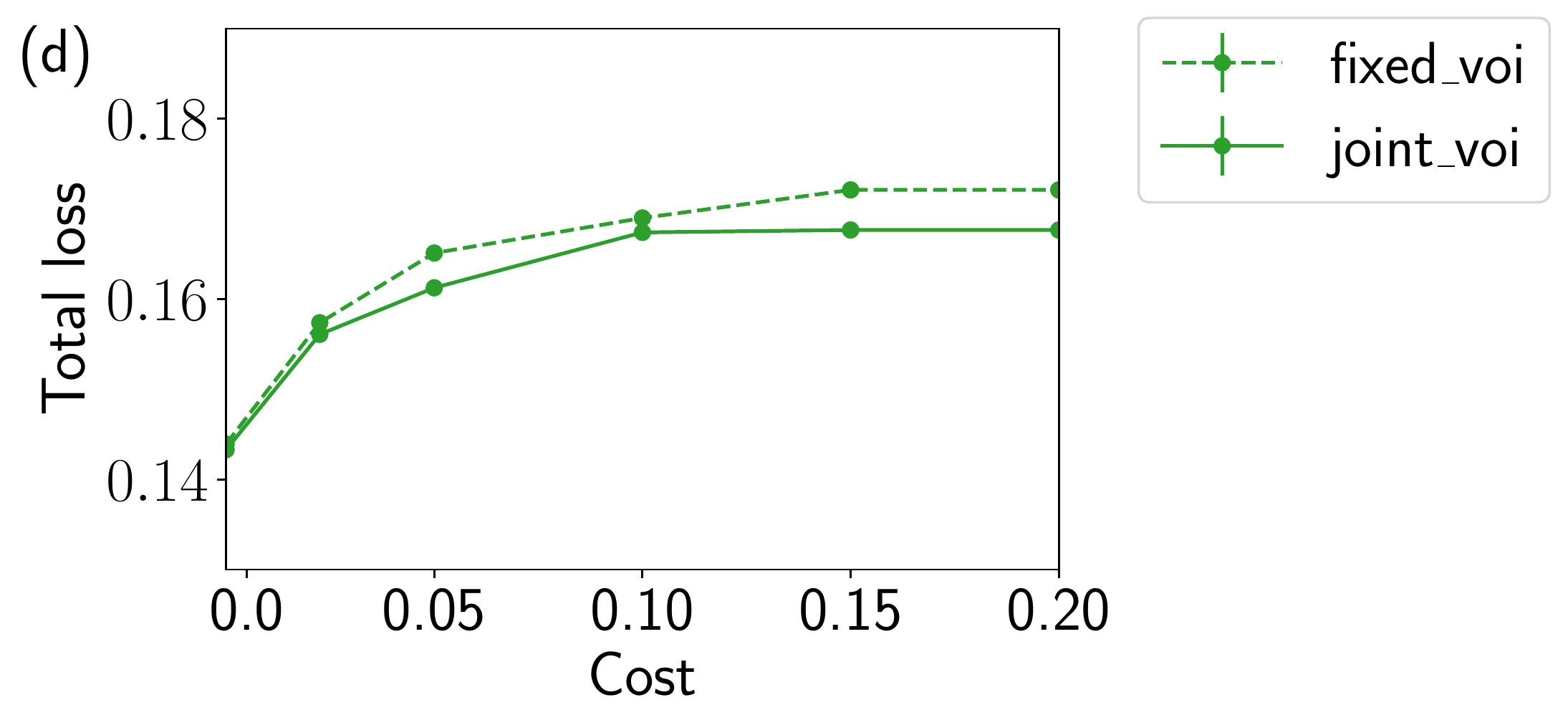}

    \caption{Total loss (classification error + cost of queries to human) as a function of the cost of a human query. Top row: All approaches. Bottom row: Zooming in on decision-theoretic approaches. (a) Galaxy Zoo  (b) CAMELYON16 (c) CAMELYON16, doubling the cost of false negatives. (d) CAMELYON 16, reducing hidden layers to 20 neurons (from 50). We omit the ``human only" baseline for Galaxy Zoo since it has over twice the loss of any other method. All differences between fixed and joint models are statistically significant for Galaxy Zoo, and on the CAMELYON16 task for the discriminative models (Student t-test, $p < 10^{-3}$). Due to the small size of the CAMELYON16 dataset (127 samples), not all VOI comparisons are statistically significant, but the larger differences approach significance (e.g, $p < 0.15$ for the point with largest difference in each of Figures \ref{fig:error}(c-d)). }
    \label{fig:error}
\end{figure*}

\begin{table}[]
\caption{Comparison of joint and fixed VOI models across a range of settings. ``Layers" gives the number of layers used in the predictive models, ``Hidden," the number of hidden units, and ``\% diff.," the percentage improvement of the joint over fixed model (given as the min, average, and max improvement in loss over costs from 0 to 0.2). } \label{table:voi}
\fontsize{8pt}{10pt}\selectfont
\begin{tabular}{lccrll}
\hline
Task         & Layers & Hidden & \multicolumn{3}{c}{\% diff. (min / avg / max)} \\ \hline
GZ           & 1      & -      & \multicolumn{3}{r}{21.8 / 38.9 / 73.3}         \\
GZ           & 2      & 50     & \multicolumn{3}{r}{2.13 / 9.02 / 14.0}         \\
GZ           & 2      & 100    & \multicolumn{3}{r}{-1.05 / 8.89 / 13.5}        \\ \hline
CAM.         & 1      & -      & \multicolumn{3}{r}{-3.10 / 4.51 /10.4}         \\
CAM. (asym.) & 1      & -      & \multicolumn{3}{r}{-1.26 / 5.13 / 15.2}        \\
CAM.         & 2      & 20     & \multicolumn{3}{r}{0.30 / 1.82 / 2.65}         \\
CAM. (asym.) & 2      & 20     & \multicolumn{3}{r}{-0.80 / 1.91 / 4.85}        \\
CAM.         & 2      & 50     & \multicolumn{3}{r}{0.00 / 0.03 / 2.31}         \\
CAM. (asym.) & 2      & 50     & \multicolumn{3}{r}{-0.67 / 1.70 / 2.28}        \\ \hline
\end{tabular}
\end{table}

\subsection{Results}

We first examine the performance of these methods for the two tasks. Fig \ref{fig:error} shows each method's total loss (combining classification error and the cost of human queries). For each model, the dashed line shows the fixed version and the solid line denotes joint. For the joint models, we train the model under a range of weightings of classification loss vs query cost, and each $x$-axis point selects the version with lowest total loss for that cost. We show discriminative models with one- and two-layer networks. Because the one- and two-layer VOI models have fairly different losses (which compresses the plots), we only show two layers. Table \ref{table:voi} gives results for all VOI configurations.

\textbf{The joint models, which optimize for complementarity, uniformly outperform or tie their fixed counterparts.} For Galaxy Zoo, joint training leads to 21-73\% reduction in loss for the one-layer VOI models and 10-15\% reduction in loss for two-layer VOI. The reductions are 10-15\% and 29\% for the one and two layer discriminative models respectively.      
For CAMELYON16, joint training improves the one-layer discriminative model by up to 20\% and the one-layer VOI model by up to 10\%. For deeper models, joint training ties the fixed approach or makes modest improvements (around 2\% reduction in loss).  
\begin{figure*}[h]
    \centering
    
    \includegraphics[width=2.25in]{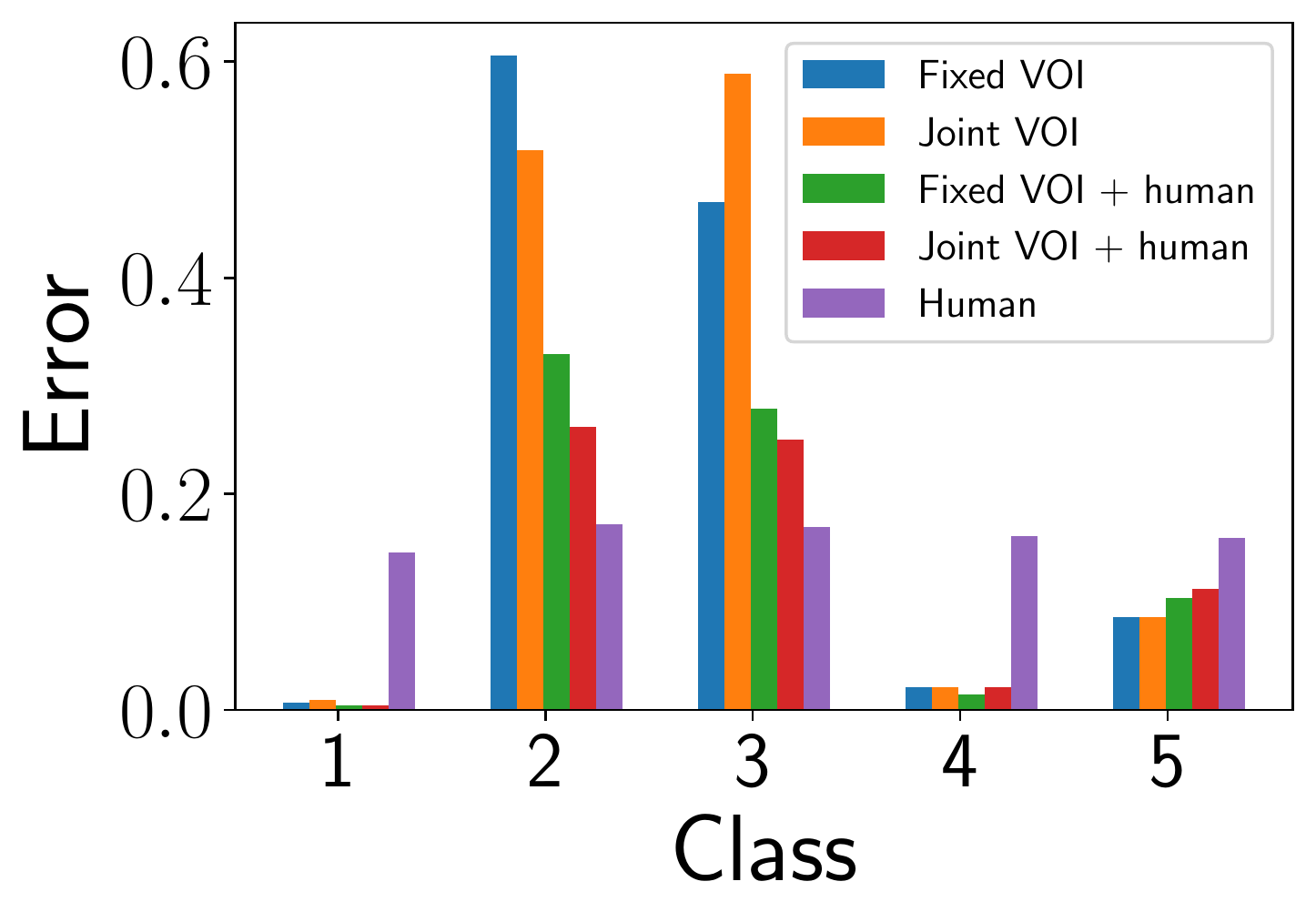}
    \includegraphics[width=2.25in]{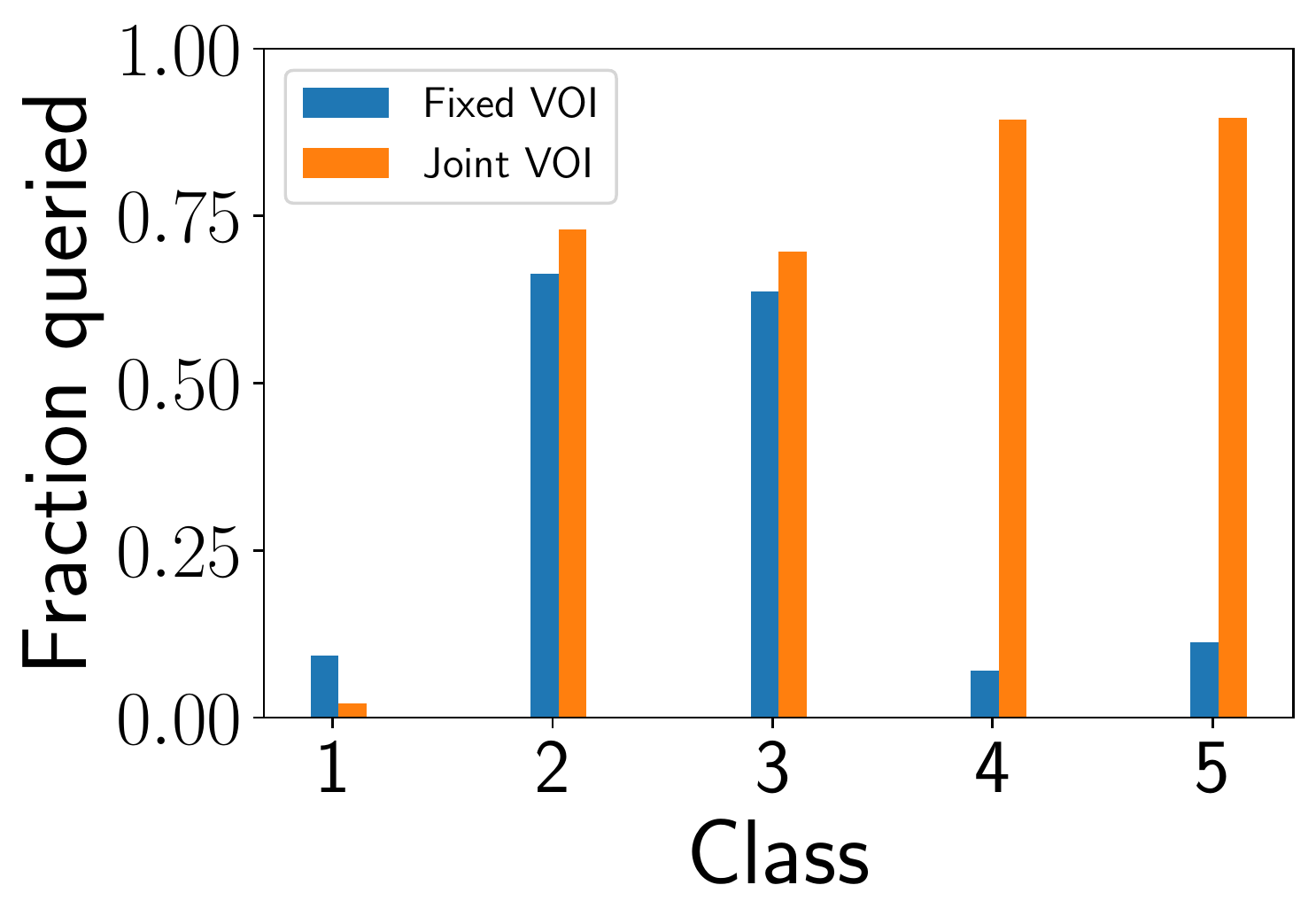}
    \caption{Detailed analysis on Galaxy Zoo task. Left: Error rate of machine versus human models for each class. Right: Fraction of instances in each class queried by the machine.}
    \label{fig:error-gz}
\end{figure*}
Next, we vary the problem setting to explore the factors that influence the benefits of joint training. First, we vary the capacity of the models, as measured by the number of hidden units. Figures \ref{fig:error}b and \ref{fig:error}d compare the total loss of different approaches when hidden unit sizes is reduced from 50 to 20. Table \ref{table:voi} examines the effect of model capacity on the VOI-based approaches.  
\textbf{Overall, joint training provides larger benefits with limited model capacity}. 
For example, for CAMELYON16, the reduction of loss from joint training for discriminative approaches is up to 15\% when hidden units are reduced to 20, whereas for the 50 neuron condition the two discriminative approaches are tied (two-layer models). 
This dovetails with earlier results that showed larger gains for shallower models. Essentially, a lower-capacity model has more potential bias (since it represents less complex hypotheses which cannot fit the ground truth as closely). This makes aligning the training process with team performance more important because some errors are inevitable; joint training helps the model focus its limited predictive ability on the most important regions. In theory, sufficiently large datasets would let us train arbitrarily complicated models that perfectly recover the ground truth, rendering simple models unnecessary. In practice, limited data requires us to prevent overfitting by restricting model capacity; maximizing the performance of simple models is valuable in many tasks. 

The second experimental modification introduces an asymmetric loss for CAMELYON16: motivated by high cost of missing diagnoses in many areas of medicine (such as failing to recognize the recurrence of illness in patients with a history of cancer), we weight false negatives twice as heavily as false positives. \textbf{The gaps between the fixed and joint models grow under asymmetric costs}. For example, in Figure \ref{fig:error}(b) (equal costs), the two-layer model performance of discriminative or VOI approaches were previously tied. In Figure \ref{fig:error}(c) (asymmetric costs), the joint approaches now outperform their fixed counterparts by up to 10\% (discriminative family) and 4.8\% (VOI). Optimizing combined team performance is especially helpful when it is necessary to prioritize between potential errors.

Finally, we examine how joint training influences the capabilities of the ML system in relation to those of humans. We start with the Galaxy Zoo task (two-layer models, 50 hidden units, cost = 0.1). Figure \ref{fig:error-gz} shows the error rates of the fixed and joint VOI models for each of the five classes when acting alone and when paired with people. Both the error rates of the two approaches on classes 2 and 3, and the way they query humans show differences, indicating that joint optimization changes how the ML system learns and makes decisions. 
The joint approach makes more queries to humans for classes that are hard for the machine and less for class 1, which is easy for the machine (note that class 1 accounts for over 70\% of instances). This behavior improves team performance on classes 2 and 3 without diminishing performance on class 1. For class 3, the error rate of the joint VOI model is higher than its counterpart when acting alone, but lower when combined with the human, a reduction in loss that cannot be simply explained by the marginal increase in human queries. This shows that the joint model can harness human input more effectively by discovering input spaces within individual classes where the benefits of complementarity can be realized, and also that 
joint training encourages the model to manage tradeoffs in accuracy to leverage the ability to query the human.
\begin{figure}
    \centering
    \includegraphics[width=2.1in]{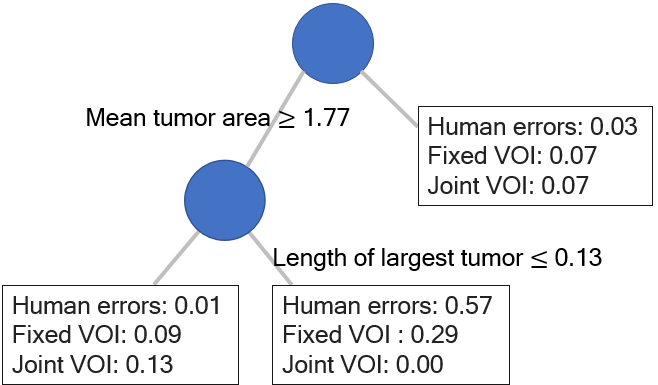}
    \caption{Error rates of humans and decision-theoretic approaches for prominent feature regions of CAMELYON16.}
    \label{fig:tree}
\end{figure}

We observe similar behavior for CAMELYON16. Here, we find clear structure in the human errors, uncovered by fitting the decision tree shown in Figure \ref{fig:tree} (for the uniform-cost task with two-layer models and 50 hidden units). Over 68\% of human errors are concentrated in a region containing just 10\% of instances, identified using two features. For each leaf, we show the error rate of the human, the fixed VOI model, and the joint VOI model. The joint model prioritizes the region that contains most of the human errors, improving from the 0.29 error rate of the fixed model to perfect accuracy. This comes at the cost of increased errors in the far-left leaf; however, in this region the human is almost perfectly accurate. Overall, this tradeoff made by the joint optimization leads to a $2\%$ overall reduction in loss. In other words, the distribution of errors incurred by the joint model shifts to complement the strengths and weaknesses of the human. 

\section{Conclusion and Future Work}

We studied how ML systems can be optimized to complement humans via the use of discriminative and decision-theoretic modeling methodologies. We evaluated the proposed approaches by performing experiments with two real-world tasks and analyzed the problem characteristics that lead to higher benefits from training focused on leveraging human-machine complementarity. The methods presented are aimed at optimizing the expected value of human-machine teamwork by responding to the shortcomings of ML systems, as well as the capabilities and blind spots of humans. With this framing, we explored the relationship between model capacity, asymmetric costs and ML-human complementarity. We see opportunities for studying additional aspects of human-machine complementarity across different settings. Directions include optimization of team performance when interactions between humans and machines extend beyond querying people for answers, such as settings with more complex, interleaved interactions and with different levels of human initiative and machine autonomy. We hope that the methods and results presented will stimulate further pursuit of opportunities for leveraging the complementarity of people and machines. 

\section*{Acknowledgments}
We thank Besmira Nushi for advice on characterizing error regions and insightful conversations throughout, as well as the CAMELYON team for providing data on pathologist panel responses.

\end{document}